\documentclass[a4paper,conference]{IEEEtran}

\usepackage[dvipsnames,svgnames,x11names]{xcolor}
\usepackage{tikzextern}
\usepackage{pgffor}

\newcommand{\extfig}[2]{\includegraphics{fig/extern/#1}}
\newcommand{\noextfig}[1]{!!!}
\newcommand{\extdata}[1]{}

\makeatletter
\renewcommand\paragraph{\@startsection{paragraph}{4}{\z@}{1ex}{-1em}{\normalfont\normalsize\bfseries}}
\makeatother

\usepackage{times}
\usepackage{epsfig}
\usepackage{graphicx}
\usepackage{amsmath}
\usepackage{amssymb}
\usepackage{multirow}
\usepackage{booktabs}
\usepackage{stfloats}
\usepackage{cite}
\usepackage{tabularx}
\usepackage{comment}
\usepackage{xspace}
\usepackage{hyperref}

\begin{document}
%
\title{Local Propagation for Few-Shot Learning}

\author{\IEEEauthorblockN{Yann Lifchitz}
\IEEEauthorblockA{Inria, Univ Rennes, CNRS, IRISA\\Safran}
\and
\IEEEauthorblockN{Yannis Avrithis}
\IEEEauthorblockA{Inria, Univ Rennes, CNRS, IRISA}
\and
\IEEEauthorblockN{Sylvaine Picard}
\IEEEauthorblockA{Safran}}


\newcommand{\head}[1]{{\smallskip\noindent\textbf{#1}}}
\newcommand{\alert}[1]{{\color{red}{#1}}}
\newcommand{\eq}[1]{(\ref{eq:#1})}

\newcommand{\Th}[1]{\textsc{#1}}
\newcommand{\mr}[2]{\multirow{#1}{*}{#2}}
\newcommand{\mc}[2]{\multicolumn{#1}{c}{#2}}
\newcommand{\tb}[1]{\textbf{#1}}
\newcommand{\ch}{\checkmark}

\newcommand{\red}[1]{{\color{red}{#1}}}
\newcommand{\blue}[1]{{\color{blue}{#1}}}
\newcommand{\green}[1]{{\color{green}{#1}}}
\newcommand{\gray}[1]{{\color{gray}{#1}}}

\newcommand{\citeme}[1]{\red{[XX]}}
\newcommand{\refme}[1]{\red{(XX)}}

\newcommand{\fig}[2][1]{\includegraphics[width=#1\columnwidth]{fig/#2}}
\newcommand{\figh}[2][1]{\includegraphics[height=#1\columnwidth]{fig/#2}}


\newcommand{\tran}{^\top}
\newcommand{\mtran}{^{-\top}}
\newcommand{\zcol}{\mathbf{0}}
\newcommand{\zrow}{\zcol\tran}

\newcommand{\ind}{\mathbbm{1}}
\newcommand{\expect}{\mathbb{E}}
\newcommand{\nat}{\mathbb{N}}
\newcommand{\zahl}{\mathbb{Z}}
\newcommand{\real}{\mathbb{R}}
\newcommand{\proj}{\mathbb{P}}
\newcommand{\prob}{\mathbf{Pr}}
\newcommand{\normal}{\mathcal{N}}

\newcommand{\mif}{\textrm{if}\ }
\newcommand{\other}{\textrm{otherwise}}
\newcommand{\minimize}{\textrm{minimize}\ }
\newcommand{\maximize}{\textrm{maximize}\ }
\newcommand{\st}{\textrm{subject\ to}\ }

\newcommand{\id}{\operatorname{id}}
\newcommand{\const}{\operatorname{const}}
\newcommand{\sgn}{\operatorname{sgn}}
\newcommand{\var}{\operatorname{Var}}
\newcommand{\mean}{\operatorname{mean}}
\newcommand{\trace}{\operatorname{tr}}
\newcommand{\diag}{\operatorname{diag}}
\newcommand{\vect}{\operatorname{vec}}
\newcommand{\cov}{\operatorname{cov}}
\newcommand{\sign}{\operatorname{sign}}
\newcommand{\prj}{\operatorname{proj}}

\newcommand{\softmax}{\operatorname{softmax}}
\newcommand{\clip}{\operatorname{clip}}

\newcommand{\defn}{\mathrel{:=}}
\newcommand{\peq}{\mathrel{+\!=}}
\newcommand{\meq}{\mathrel{-\!=}}

\newcommand{\floor}[1]{\left\lfloor{#1}\right\rfloor}
\newcommand{\ceil}[1]{\left\lceil{#1}\right\rceil}
\newcommand{\inner}[1]{\left\langle{#1}\right\rangle}
\newcommand{\norm}[1]{\left\|{#1}\right\|}
\newcommand{\abs}[1]{\left|{#1}\right|}
\newcommand{\frob}[1]{\norm{#1}_F}
\newcommand{\card}[1]{\left|{#1}\right|\xspace}
\newcommand{\diff}{\mathrm{d}}
\newcommand{\der}[3][]{\frac{d^{#1}#2}{d#3^{#1}}}
\newcommand{\pder}[3][]{\frac{\partial^{#1}{#2}}{\partial{#3^{#1}}}}
\newcommand{\ipder}[3][]{\partial^{#1}{#2}/\partial{#3^{#1}}}
\newcommand{\dder}[3]{\frac{\partial^2{#1}}{\partial{#2}\partial{#3}}}

\newcommand{\wb}[1]{\overline{#1}}
\newcommand{\wt}[1]{\widetilde{#1}}

\def\xssp{\hspace{1pt}}
\def\ssp{\hspace{3pt}}
\def\msp{\hspace{5pt}}
\def\lsp{\hspace{12pt}}

\newcommand{\cA}{\mathcal{A}}
\newcommand{\cB}{\mathcal{B}}
\newcommand{\cC}{\mathcal{C}}
\newcommand{\cD}{\mathcal{D}}
\newcommand{\cE}{\mathcal{E}}
\newcommand{\cF}{\mathcal{F}}
\newcommand{\cG}{\mathcal{G}}
\newcommand{\cH}{\mathcal{H}}
\newcommand{\cI}{\mathcal{I}}
\newcommand{\cJ}{\mathcal{J}}
\newcommand{\cK}{\mathcal{K}}
\newcommand{\cL}{\mathcal{L}}
\newcommand{\cM}{\mathcal{M}}
\newcommand{\cN}{\mathcal{N}}
\newcommand{\cO}{\mathcal{O}}
\newcommand{\cP}{\mathcal{P}}
\newcommand{\cQ}{\mathcal{Q}}
\newcommand{\cR}{\mathcal{R}}
\newcommand{\cS}{\mathcal{S}}
\newcommand{\cT}{\mathcal{T}}
\newcommand{\cU}{\mathcal{U}}
\newcommand{\cV}{\mathcal{V}}
\newcommand{\cW}{\mathcal{W}}
\newcommand{\cX}{\mathcal{X}}
\newcommand{\cY}{\mathcal{Y}}
\newcommand{\cZ}{\mathcal{Z}}

\newcommand{\vA}{\mathbf{A}}
\newcommand{\vB}{\mathbf{B}}
\newcommand{\vC}{\mathbf{C}}
\newcommand{\vD}{\mathbf{D}}
\newcommand{\vE}{\mathbf{E}}
\newcommand{\vF}{\mathbf{F}}
\newcommand{\vG}{\mathbf{G}}
\newcommand{\vH}{\mathbf{H}}
\newcommand{\vI}{\mathbf{I}}
\newcommand{\vJ}{\mathbf{J}}
\newcommand{\vK}{\mathbf{K}}
\newcommand{\vL}{\mathbf{L}}
\newcommand{\vM}{\mathbf{M}}
\newcommand{\vN}{\mathbf{N}}
\newcommand{\vO}{\mathbf{O}}
\newcommand{\vP}{\mathbf{P}}
\newcommand{\vQ}{\mathbf{Q}}
\newcommand{\vR}{\mathbf{R}}
\newcommand{\vS}{\mathbf{S}}
\newcommand{\vT}{\mathbf{T}}
\newcommand{\vU}{\mathbf{U}}
\newcommand{\vV}{\mathbf{V}}
\newcommand{\vW}{\mathbf{W}}
\newcommand{\vX}{\mathbf{X}}
\newcommand{\vY}{\mathbf{Y}}
\newcommand{\vZ}{\mathbf{Z}}

\newcommand{\va}{\mathbf{a}}
\newcommand{\vb}{\mathbf{b}}
\newcommand{\vc}{\mathbf{c}}
\newcommand{\vd}{\mathbf{d}}
\newcommand{\ve}{\mathbf{e}}
\newcommand{\vf}{\mathbf{f}}
\newcommand{\vg}{\mathbf{g}}
\newcommand{\vh}{\mathbf{h}}
\newcommand{\vi}{\mathbf{i}}
\newcommand{\vj}{\mathbf{j}}
\newcommand{\vk}{\mathbf{k}}
\newcommand{\vl}{\mathbf{l}}
\newcommand{\vm}{\mathbf{m}}
\newcommand{\vn}{\mathbf{n}}
\newcommand{\vo}{\mathbf{o}}
\newcommand{\vp}{\mathbf{p}}
\newcommand{\vq}{\mathbf{q}}
\newcommand{\vr}{\mathbf{r}}
\newcommand{\Vs}{\mathbf{s}}
\newcommand{\vt}{\mathbf{t}}
\newcommand{\vu}{\mathbf{u}}
\newcommand{\vv}{\mathbf{v}}
\newcommand{\vw}{\mathbf{w}}
\newcommand{\vx}{\mathbf{x}}
\newcommand{\vy}{\mathbf{y}}
\newcommand{\vz}{\mathbf{z}}

\newcommand{\vone}{\mathbf{1}}
\newcommand{\vzero}{\mathbf{0}}

\newcommand{\valpha}{{\boldsymbol{\alpha}}}
\newcommand{\vbeta}{{\boldsymbol{\beta}}}
\newcommand{\vgamma}{{\boldsymbol{\gamma}}}
\newcommand{\vdelta}{{\boldsymbol{\delta}}}
\newcommand{\vepsilon}{{\boldsymbol{\epsilon}}}
\newcommand{\vzeta}{{\boldsymbol{\zeta}}}
\newcommand{\veta}{{\boldsymbol{\eta}}}
\newcommand{\vtheta}{{\boldsymbol{\theta}}}
\newcommand{\viota}{{\boldsymbol{\iota}}}
\newcommand{\vkappa}{{\boldsymbol{\kappa}}}
\newcommand{\vlambda}{{\boldsymbol{\lambda}}}
\newcommand{\vmu}{{\boldsymbol{\mu}}}
\newcommand{\vnu}{{\boldsymbol{\nu}}}
\newcommand{\vxi}{{\boldsymbol{\xi}}}
\newcommand{\vomikron}{{\boldsymbol{\omikron}}}
\newcommand{\vpi}{{\boldsymbol{\pi}}}
\newcommand{\vrho}{{\boldsymbol{\rho}}}
\newcommand{\vsigma}{{\boldsymbol{\sigma}}}
\newcommand{\vtau}{{\boldsymbol{\tau}}}
\newcommand{\vupsilon}{{\boldsymbol{\upsilon}}}
\newcommand{\vphi}{{\boldsymbol{\phi}}}
\newcommand{\vchi}{{\boldsymbol{\chi}}}
\newcommand{\vpsi}{{\boldsymbol{\psi}}}
\newcommand{\vomega}{{\boldsymbol{\omega}}}

\newcommand{\rLambda}{\mathrm{\Lambda}}
\newcommand{\rSigma}{\mathrm{\Sigma}}

\newcommand{\vLambda}{\bm{\rLambda}}
\newcommand{\vSigma}{\bm{\rSigma}}

\makeatletter
\newcommand*\bdot{\mathpalette\bdot@{.7}}
\newcommand*\bdot@[2]{\mathbin{\vcenter{\hbox{\scalebox{#2}{$\m@th#1\bullet$}}}}}
\makeatother

\makeatletter
\DeclareRobustCommand\onedot{\futurelet\@let@token\@onedot}
\def\@onedot{\ifx\@let@token.\else.\null\fi\xspace}

\def\eg{\emph{e.g}\onedot} \def\Eg{\emph{E.g}\onedot}
\def\ie{\emph{i.e}\onedot} \def\Ie{\emph{I.e}\onedot}
\def\cf{\emph{cf}\onedot} \def\Cf{\emph{Cf}\onedot}
\def\etc{\emph{etc}\onedot} \def\vs{\emph{vs}\onedot}
\def\wrt{w.r.t\onedot} \def\dof{d.o.f\onedot} \def\aka{a.k.a\onedot}
\def\etal{\emph{et al}\onedot}
\makeatother

\newcommand{\kay}{k}


\newcommand{\base}[1]{\dot{#1}}

\newcommand{\bC}{\base{C}}
\newcommand{\bD}{\base{D}}
\newcommand{\bF}{\base{F}}
\newcommand{\bN}{\base{N}}
\newcommand{\bU}{\base{U}}
\newcommand{\bW}{\base{W}}
\newcommand{\bX}{\base{X}}
\newcommand{\bY}{\base{Y}}

\newcommand{\bb}{\base{b}}
\newcommand{\bc}{\base{c}}
\newcommand{\bk}{\base{k}}
\newcommand{\bn}{\base{n}}
\newcommand{\bx}{\base{x}}
\newcommand{\by}{\base{y}}

\newcommand{\bva}{\base{\va}}
\newcommand{\bvb}{\base{\vb}}
\newcommand{\bvp}{\base{\vp}}
\newcommand{\bvq}{\base{\vq}}
\newcommand{\bvw}{\base{\vw}}
\newcommand{\bvx}{\base{\vx}}
\newcommand{\bvy}{\base{\vy}}

\newcommand{\bell}{\base{\ell}}
\newcommand{\btheta}{\base{\theta}}


\newcommand{\new}[1]{#1}

\newcommand{\nC}{\new{C}}
\newcommand{\nD}{\new{D}}
\newcommand{\nF}{\new{F}}
\newcommand{\nN}{\new{N}}
\newcommand{\nU}{\new{U}}
\newcommand{\nW}{\new{W}}
\newcommand{\nX}{\new{X}}
\newcommand{\nY}{\new{Y}}

\newcommand{\nb}{\new{b}}
\newcommand{\nc}{\new{c}}
\newcommand{\nf}{\new{f}}
\newcommand{\nk}{s}
\newcommand{\nn}{\new{n}}
\newcommand{\nx}{\new{x}}
\newcommand{\ny}{\new{y}}

\newcommand{\nva}{\new{\va}}
\newcommand{\nvb}{\new{\vb}}
\newcommand{\nvp}{\new{\vp}}
\newcommand{\nvq}{\new{\vq}}
\newcommand{\nvw}{\new{\vw}}
\newcommand{\nvx}{\new{\vx}}
\newcommand{\nvy}{\new{\vy}}

\newcommand{\nell}{\new{\ell}}
\newcommand{\ntheta}{\new{\theta}}


\newcommand{\pre}[1]{#1^\circ}

\newcommand{\pC}{\pre{C}}
\newcommand{\pD}{\pre{D}}
\newcommand{\pF}{\pre{F}}
\newcommand{\pN}{\pre{N}}
\newcommand{\pU}{\pre{U}}
\newcommand{\pW}{\pre{W}}
\newcommand{\pX}{\pre{X}}
\newcommand{\pY}{\pre{Y}}

\newcommand{\pb}{\pre{b}}
\newcommand{\pc}{\pre{c}}
\newcommand{\pf}{\pre{f}}
\newcommand{\pk}{\pre{k}}
\newcommand{\pn}{\pre{n}}
\newcommand{\py}{\pre{y}}

\newcommand{\pva}{\pre{\va}}
\newcommand{\pvb}{\pre{\vb}}
\newcommand{\pvp}{\pre{\vp}}
\newcommand{\pvq}{\pre{\vq}}
\newcommand{\pvw}{\pre{\vw}}
\newcommand{\pvx}{\pre{\vx}}
\newcommand{\pvy}{\pre{\vy}}

\newcommand{\pell}{\pre{\ell}}
\newcommand{\ptheta}{\pre{\theta}}

\pgfplotsset{every axis plot/.append style={mark=*}}
\tikzstyle{cub} = [blue]
\tikzstyle{mi} = [red]
\tikzstyle{five} = [dashed]
\tikzstyle{tran} = [mark=o,opacity=.5]

\pgfplotstableread{
	queries    CUB1    CUB5     MI1     MI5   GCUB5   ANBNN    NBNN     GAP    AGAP  nACUB5 nAGCUB5
  	      1  0.7932  0.9152  0.6443  0.8026  0.8878  0.9140  0.8959  0.9038  0.9124  0.9090  0.8796
	      5  0.8502  0.9233  0.6959  0.8145  0.9084  0.9140  0.8959  0.9038  0.9124  0.9124  0.8975
	     10  0.8718  0.9312  0.7149  0.8247  0.9186  0.9140  0.8959  0.9038  0.9124  0.9185  0.9056
	     15  0.8777  0.9328  0.7239  0.8276  0.9188  0.9140  0.8959  0.9038  0.9124  0.9201  0.9063
	     20  0.8802  0.9369  0.7259  0.8334  0.9221  0.9140  0.8959  0.9038  0.9124  0.9231  0.9076
}{\univ}

\pgfplotstableread{
	thres    GAP1    GAP5   MGAP1   MGAP5
	    0  0.7876  0.9184  0.6519  0.8308
	  0.1  0.7896  0.9196  0.6544  0.8329
	  0.2  0.7986  0.9231  0.6604  0.8362
	  0.3  0.8067  0.9253  0.6648  0.8385
	  0.4  0.8094  0.9251  0.6674  0.8394
	  0.5  0.8054  0.9224  0.6643  0.8360
	  0.6  0.7892  0.9144  0.6550  0.8274
}{\attgap}

\pgfplotstableread{
	thres    CUB1    CUB5   TCUB1   TCUB5     MI1     MI5    TMI1    TMI5
	    0  0.8156  0.9237  0.8766  0.9346  0.6697  0.8167  0.7256  0.8318
	  0.1  0.8178  0.9248  0.8791  0.9353  0.6709  0.8188  0.7283  0.8340
	  0.2  0.8268  0.9290  0.8893  0.9393  0.6754  0.8250  0.7405  0.8416
	  0.3  0.8302  0.9334  0.8989  0.9441  0.6781  0.8293  0.7506  0.8477
	  0.4  0.8282  0.9348  0.9032  0.9451  0.6780  0.8316  0.7553  0.8511
	  0.5  0.8205  0.9342  0.9032  0.9444  0.6738  0.8311  0.7546  0.8505
	  0.6  0.8128  0.9330  0.9020  0.9426  0.6684  0.8287  0.7530  0.8483
}{\attllp}

\pgfplotstableread{
	clust    CUB1    CUB5   TCUB1   TCUB5     MI1     MI5    TMI1    TMI5
	   10  0.8048  0.9107  0.8854  0.9297  0.6542  0.8144  0.7097  0.8121
	   20  0.8220  0.9242  0.8981  0.9398  0.6704  0.8256  0.7390  0.8345
	   30  0.8281  0.9291  0.8995  0.9427  0.6769  0.8290  0.7466  0.8421
	   40  0.8314  0.9316  0.8985  0.9434  0.6792  0.8297  0.7500  0.8454
	   50  0.8312  0.9326  0.8951  0.9441  0.6791  0.8286  0.7508  0.8472
	   60  0.8302  0.9334  0.8917  0.9438  0.6781  0.8267  0.7506  0.8477
	   70  0.8265  0.9334  0.8812  0.9435  0.6763  0.8238  0.7495  0.8476
}{\pool}

\maketitle

\begin{abstract}
The challenge in few-shot learning is that available data is not enough to capture the underlying distribution. To mitigate this, two emerging directions are (a) using local image representations, essentially multiplying the amount of data by a constant factor, and (b) using more unlabeled data, for instance by transductive inference, jointly on a number of queries. In this work, we bring these two ideas together, introducing \emph{local propagation}. We treat local image features as independent examples, we build a graph on them and we use it to propagate both the features themselves and the labels, known and unknown. Interestingly, since there is a number of features per image, even a single query gives rise to transductive inference. As a result, we provide a universally safe choice for few-shot inference under both non-transductive and transductive settings, improving accuracy over corresponding methods. This is in contrast to existing solutions, where one needs to choose the method depending on the quantity of available data.
\end{abstract}

\IEEEpeerreviewmaketitle

\section{Introduction}
\label{sec:intro}

\emph{Few-shot learning}~\cite{vinyals2016,snell2017,wang2018} is the problem of learning new tasks from few examples, possibly transferring knowledge from previous tasks. Against the mainstream paradigm of having lots of labeled data in deep learning, it limits not only the amount of supervision but also the amount of data. Given the variability of appearance in few-shot image classification benchmarks, learning from few examples without knowledge of the underlying distribution is truly challenging.

Few-shot learning has been little studied before deep learning~\cite{BaUl05}. Research on few-shot learning is recently becoming very popular, but is not very mature. On one hand, it is often connected to \emph{meta-learning}~\cite{vilalta2002perspective} in the sense of learning to compare~\cite{vinyals2016} or to optimize~\cite{finn2017}, giving rise to complex ideas involving second-order derivatives. On the other hand, it boils down to \emph{representation learning}, using \eg metric learning~\cite{li2019revisiting} or parametric classifiers~\cite{chen2019}, followed by nearest neighbor classifiers at inference~\cite{wang2019simpleshot}.

While there are several approaches on \emph{generating} more data~\cite{wang2018,Wang_2020_CVPR}, global spatial pooling into compact image representations ignores the rich data that is hidden in each given example. Each image is inherently a collection of data, which has been exploited by \emph{dense classification} (DC) at representation learning~\cite{Lifchitz19} and \emph{na\"ive Bayes nearest neighbor} (NBNN)~\cite{LiWang_2019_CVPR} at inference.

Using \emph{unlabeled data} is another popular direction of research, leveraging existing results from transductive inference~\cite{LLP+19}, semi-supervised learning~\cite{GaBr18} and self-supervised representation learning~\cite{gidaris2019boosting}. \emph{Graph-based methods} are at the core of this effort, using for instance label propagation~\cite{LLP+19}, feature propagation~\cite{rodriguez2020embedding} and graph neural networks~\cite{GaBr18}.

This work is an attempt to bridge these two ideas, \ie, local representations~\cite{LiWang_2019_CVPR} and propagation~\cite{rodriguez2020embedding}, into a common framework. Essentially, NBNN~\cite{LiWang_2019_CVPR} measures the \emph{average similarity} of local representations of a given image to local representations of all images in a class; while feature or label propagation~\cite{rodriguez2020embedding} replaces raw (Euclidean) similarities with similarities taking into account the manifold structure of the data distribution, and measures a single \emph{manifold similarity} of a given image to a class. Our \emph{local propagation} combines both by measuring the \emph{average manifold similarity} of local representations of a given image to local representations of all images in a class.

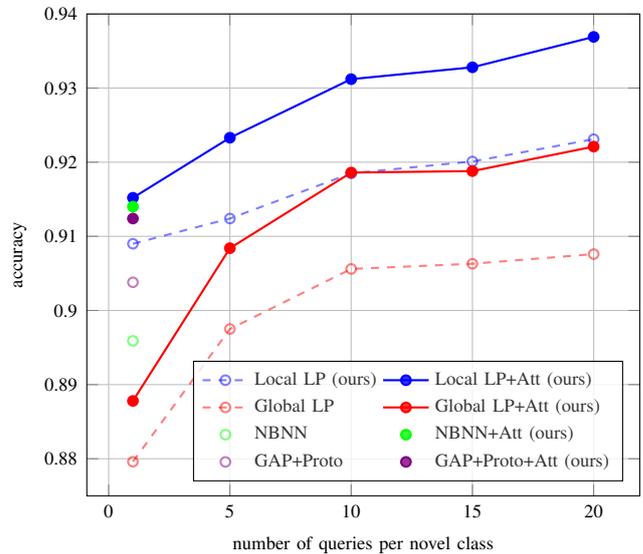
\begin{figure}
\centering
\extfig{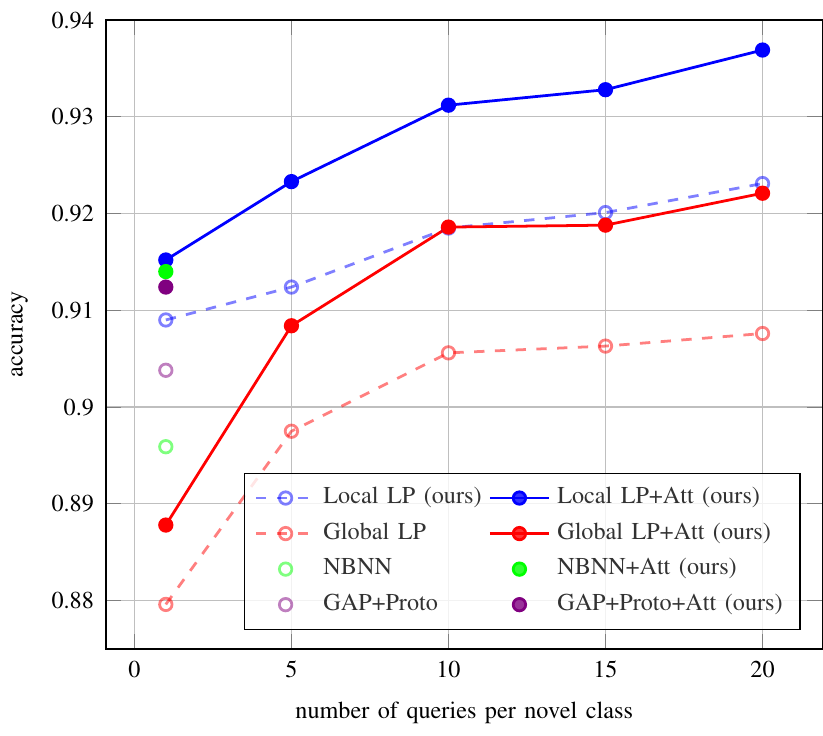}{
\begin{tikzpicture}
\begin{axis}[
	width=\linewidth,
	height=0.90\linewidth,
	font=\scriptsize,
	xlabel={number of queries per novel class},
	ylabel={accuracy},
	ymin = .8750,
	ymax = .94,
	legend pos=south east,
	legend columns=2,
]

	\addplot[blue,dash] table[x=queries,y=nACUB5] \univ; \leg{Local LP (ours)};
	\addplot[blue] table[x=queries,y=CUB5]        \univ; \leg{Local LP+Att (ours)};
  	\addplot[red,dash] table[x=queries,y=nAGCUB5] \univ; \leg{Global LP};
  	\addplot[red] table[x=queries,y=GCUB5]        \univ; \leg{Global LP+Att (ours)};

  	\addplot[green, only marks,dash] coordinates {(1,0.8959)}; \leg{NBNN};
  	\addplot[green, only marks]      coordinates {(1,0.9140)}; \leg{NBNN+Att (ours)};
  	\addplot[violet,only marks,dash] coordinates {(1,0.9038)}; \leg{GAP+Proto};
  	\addplot[violet,only marks]      coordinates {(1,0.9124)}; \leg{GAP+Proto+Att (ours)};

\end{axis}
\end{tikzpicture}
}
\caption{CUB 5-way 5-shot classification accuracy \vs number of queries per novel class. Our local label propagation (LP) outperforms transductive and non-transductive baselines in all settings. By contrast, global LP only competes with non-trasductive methods when at least 10 unlabelled queries are available. Spatial attention (Att) is also our contribution. We use feature propagation~\cite{rodriguez2020embedding} for all methods. We use feature pooling for local propagation. Please see \autoref{sec:exp} for more details.}
\label{fig:univ}
\end{figure}

Concretely, we learn a representation using DC~\cite{Lifchitz19} and we apply local propagation at inference, without meta-learning: We break down the convolutional activations of support and query images into pieces corresponding to different spatial positions, consider all these pieces as different examples, and then apply feature or label propagation~\cite{rodriguez2020embedding} to these examples. Pieces originating in support examples inherit their labels as in DC~\cite{Lifchitz19} and NBNN~\cite{LiWang_2019_CVPR}, while pieces originating in queries are unlabeled. Since there are a number of unlabeled pieces per image, this gives rise to \emph{transductive inference} even in the case of a single query image. As shown in \autoref{fig:univ}, this means that our method is a \emph{universally safe choice} regardless of the amount of available unlabeled data.

In summary, we make the following contributions:

\begin{itemize}
	\item We study graph-based propagation on \emph{local} (pixel) and \emph{semi-local} (clusters) representations across of images for \emph{feature and label propagation} for the first time.
	\item We apply this idea to few-shot learning, effectively \emph{generating} more data and \emph{propagating} through it, bringing even the case of single queries into \emph{transductive inference}.
	\item We show that an extremely simple \emph{spatial attention} mechanism is not only essential in our local propagation, but also brings significant gains in all baselines.
	\item We show consistent gains in most datasets and settings, including transductive and non-transductive.
\end{itemize}

\section{Related work}
\label{sec:related}

\head{Few shot learning.}
While studies before deep learning have been scarce~\cite{MiMV00,fei2003bayesian,BaUl05}, few-shot learning has become a very popular problem beginning mainly from \emph{matching networks}~\cite{vinyals2016} and \emph{prototypical networks}~\cite{snell2017}. Seen as a \emph{meta-learning} problem of learning to compare in episodes, these solutions amount essentially to \emph{metric learning}, and indeed such methods have been revisited in the context of few-shot learning~\cite{wu2018improving,li2019revisiting}. Simpler methods have highlighted the importance of \emph{representation learning}. These include for instance nearest-neighbor classifiers without meta-learning~\cite{wang2019simpleshot} and simple variants of supervised classifiers like a \emph{cosine classifier}~\cite{qi2018,gidaris2018,chen2019}.

\head{More data.}
Since the main challenge in few-shot learning is the lack of data, several approaches focus on finding more. These include \emph{augmentation} in the feature space~\cite{mangla2020charting} or by combining spatial elements of images~\cite{seo2020self,le2020continual}, \emph{generation} in the feature space~\cite{wang2018} or images~\cite{chen2018semantic,ZCG+18}, \emph{image-to-image translation}~\cite{liu2019few,Wang_2020_CVPR}, using \emph{base-class data}~\cite{Li_2019_ICCV,afrasiyabi2019associative}, or even \emph{true additional data}, unlabeled~\cite{Douze_2018_CVPR} or weakly labeled~\cite{iscen2019graph}. By contrast, we generate more data ``for free'' by just looking more carefully within the existing data.

\head{Transductive inference.}
Another possibility is to consider multiple queries jointly and exploit their distribution, even though they are unlabeled. This gives rise to transductive inference~\cite{RoES13,nichol2018,LLP+19}. Most well-known are \emph{transductive propagation networks} (TPN)~\cite{LLP+19}, which use \emph{label propagation}~\cite{ZBL+03} in a meta-learning setting. Recently, this direction is becoming very popular~\cite{Kim_2019_CVPR,Qiao_2019_ICCV,hu2020exploiting,kye2020transductive,rodriguez2020embedding,wang2020instance,yang2020dpgn}. Most related to our work is the very recent \emph{embedding propagation}~\cite{rodriguez2020embedding}, which propagates the features as well as the labels. We do the same without meta-learning and most importantly, all propagation is \emph{local}.

\head{Semi/self-supervised learning.}
Using unlabeled data in an \emph{inductive setting} gives rise to \emph{semi-supervised learning} by using pseudo-labels~\cite{RRTSST18}, graph-based methods~\cite{GaBr18}, or feature-space augmentation~\cite{yu2019transmatch,mangla2020charting}. It is also common to use auxiliary unsupervised objectives like \emph{rotation}~\cite{gidaris2019boosting,mangla2020charting}. While we do not address an inductive setting, our work is a direct extension of graph-based methods, hence it can be applied to an inductive setting too, much like label propagation itself~\cite{Iscen_2019_CVPR}, or combined with any other objective.

\head{Attention.}
It is common to use attention and adaptation mechanisms in the feature space~\cite{vinyals2016,mishra2018,ren2018incremental,gidaris2018,oreshkin2018,Li_2019_CVPR}. However, despite being the subject of a pioneering work in 2005~\cite{BaUl05}, looking at local information in images has not been studied more recently in few-shot learning, until \emph{dense classification} (DC)~\cite{Lifchitz19} and \emph{na\"ive Bayes nearest neighbor} (NBNN)~\cite{LiWang_2019_CVPR}. We use the former for representation learning. The latter is similar to our work in using local representations at inference, the difference being that we apply propagation. These works have been followed by studies on \emph{spatial attention}~\cite{Wertheimer_2019_CVPR,Zhang_2019_CVPR,xv2019multi,lifchitz2020few} and \emph{alignment}~\cite{hou2019cross,Hao_2019_ICCV,Wu_2019_ICCV,zhang2020deepemd}. We experiment with an extremely simple spatial attention mechanism in this work, which requires no learning and boosts significantly all baselines.

\head{Local propagation.}
Whatever is propagated (similarities, features, or labels), there are two extremes in graph-based propagation. At one extreme, vertices are \emph{global} representations of images, and the graph represents a dataset. This can be used \eg for similarity search~\cite{ZWG+03} or semi-supervised classification~\cite{ZhGh02,ZBL+03}. At the other extreme, vertices are \emph{local} representations of pixels in an image, which can be used \eg for interactive~\cite{Grad06,KiLL08} or semantic~\cite{Bertasius_2017_CVPR} segmentation, or both~\cite{Vernaza_2017_CVPR}. \emph{Regional representations} across images have been used for similarity search~\cite{ITA+17}, but we believe we are the first to use \emph{local} (pixel) or \emph{semi-local} (clusters) representations across images for feature or label propagation.
\section{Preliminaries}
\label{sec:prelim}

\head{Problem.}
A few-shot classification task comprises a dataset $\nD \defn \{(\nvx_i, \nvy_i)\}_{i=1}^{\nn}$ of \emph{support examples} $\nvx_i \in \cX$ and \emph{labels} $\nvy_i$, where $\cX$ is an input space. Each label is represented by a \emph{one-hot} vector $\nvy_i \defn \ve_{\nell_i}$, where $\{\ve_j\}_{j=1}^{\nc}$ is the standard basis of $\real^{\nc}$, $\nell_i \in \nC$ is a \emph{label index} and $\nC \defn [\nc] \defn \{1,\dots,\nc\}$ is a set of \emph{novel (unseen) classes}. The number $\nn$ of support examples is assumed to be small. The most common setting is $\nk$ examples per novel class, with \eg $\nk \in \{1,5,10\}$, so that $\nn = \nc \nk$, referred to as \emph{$\nc$-way}, \emph{$\nk$-shot} classification. The objective of the task is to learn a classifier $f: \cX \mapsto \real_+^{\nc}$ on the support data $\nD$. This classifier maps a new \emph{query example} $\vx$ from $\cX$ to a probability distribution $\vp = f(\vx)$. A discrete \emph{prediction} $\pi(\vp)$ in $\nC$ follows, where
\begin{equation}
	\pi(\vp) \defn \arg\max_{j \in [\nc]} p_j
\label{eq:pred}
\end{equation}
is the class of of maximum probability and $p_j$ is the $j$-th element of $\vp$.

Before we are presented with few-shot classification tasks, we are given a dataset $\bD \defn \{(\bvx_i, \bvy_i)\}_{i=1}^{\bn}$ of \emph{training examples} $\bvx_i \in \cX$ and \emph{labels} $\bvy_i \defn \ve_{\bell_i}$, where $\bell_i \in \bC$ and $\bC \defn [\bc]$ is a set of \emph{base classes}, disjoint from $\nC$. The number $\bn$ of training examples is assumed to be large enough to learn a \emph{representation} of data in $\cX$, or otherwise accumulate knowledge that facilitates solving new tasks. We call this process \emph{base training}. The problem of few-shot classification amounts to designing both the base training process given $\bD$ and how to solve new tasks given $\nD$.

\head{Transductive setting.}
It is possible that in each few-shot classification task we are given a \emph{set} $Q \defn \{\vq_i\}_{i=1}^q$ of query examples and a prediction is required for all queries in $Q$. In this case, although query examples are unlabeled, we can take advantage of this additional data and learn a classifier $f$ that is a function of both the labeled support data $\nD$ and the unlabeled queries $Q$. This transductive setting implies \emph{semi-supervised learning}.

\head{Representation.}
The classifier is built on top of an \emph{embedding function} $\phi_{\theta}: \cX \to \real^{r \times d}$, with parameters $\theta$ that are learned at base training. Given an example $\vx \in \cX$, this function yields a $r \times d$ \emph{feature tensor} $\phi_{\theta}(\vx)$, where $r$ represents the dimensions of a spatial domain $\Omega$ and $d$ the feature dimensions. For $\cX$ comprising 2D images, the feature is a $w \times h \times d$ tensor that is the activation of the last convolutional layer, $r \defn w \times h$ is the spatial resolution and $\Omega \defn [w] \times [h]$ is the spatial domain. The feature can still be a vector in $\real^d$ in the special case $r=1$, \eg using \emph{global spatial pooling}. The feature tensor $F \defn \phi_{\theta}(\vx)$ contains a feature vector $F(\vr) = \phi_{\theta}(\vx)(\vr) \in \real^d$ for each spatial position $\vr \in \Omega$.

\section{Background}
\label{sec:back}

\head{Cosine classifier.}
Initially used in face verification~\cite{ranjan2017l2,wang2017normface}, a simple form of base training that was introduced to few-shot learning independently by Qi \etal~\cite{qi2018} and Gidaris and Komodakis~\cite{gidaris2018}, is to learn a parametric linear classifier that consists of a fully-connected layer without bias on top of the embedding function $\phi_{\theta}$, followed by softmax and cross-entropy. If $W \defn (\bvw_j)_{j=1}^{\bc}$ is the collection of class weights with $\bvw_j \in \real^{r \times d}$, the classifier is defined by
\begin{align}
	f_{\theta,W}(\vx) \defn \vsigma \left( \rho [\cos(\phi_{\theta}(\vx), \bvw_j)]_{j=1}^{\bc} \right),
\label{eq:cosine}
\end{align}
for $\vx \in \cX$, where $\cos$ is \emph{cosine similarity}, $\rho \in \real^+$ is a trainable \emph{scale parameter} and $\vsigma: \real^m \to \real_+^m$ is the \emph{softmax function} $\vsigma(\va) \defn (e^{a_1},\dots,e^{a_{\bc}}) / \sum_j e^{a_j}$ for $\va \in \real^{\bc}$. The representations (features and class weights) either retain resolution $r > 1$ and are \emph{flattened} to vectors of length $rd$, or are \emph{pooled} to vectors of length $d$ ($r=1$), by global spatial pooling. Base training amounts to minimizing the \emph{cost function}
\begin{align}
	J(\bD; \theta, W) \defn \sum_{i=1}^n \ell(f_{\theta,W}(\bvx_i), \bvy_i)
\label{eq:ce-cost}
\end{align}
over $\theta,W$, where $\ell(\vp,\vy) \defn -\log \inner{\vy,\vp}$ for $\vy \in \{0,1\}^{\bc}$ and $\vp \in \real_+^{\bc}$, is the \emph{cross-entropy} loss.


\head{Prototypes.}
A popular classifier for few-shot classification tasks is the \emph{prototype classifier}, introduced by Snell \etal~\cite{snell2017}. If $\cI_j \defn \{ i \in [\nn]: \nvy_i = \ve_j \}$ denotes the indices of support examples labeled in class $j$, then the \emph{prototype} of this class $j$ is given by the average features
\begin{align}
	\vmu_j \defn \frac{1}{|\cI_j|} \sum_{i \in \cI_j} \phi_{\ntheta}(\nvx_i)
\label{eq:proto}
\end{align}
of those examples for $j \in \nC$. Again, features are either flattened or pooled to vectors first. Then, denoting by $M \defn (\vmu_j)_{j=1}^{\nc}$ the collection of prototypes, a query $\vx \in \cX$ is mapped to $f_{\ntheta,M}(\vx)$, as defined by~\eq{cosine}.


\head{Na\"ive Bayes nearest neighbor (NBNN).}
In the revival~\cite{LiWang_2019_CVPR} of the classic image-to-class approach~\cite{BoSI08}, one collects, for each class $j \in [\nc]$, the features $V_j \defn \{ \phi_{\ntheta}(\nvx_i)(\vr) \}_{i \in \cI_j, \vr \in \Omega}$ of all spatial positions of all support examples labeled in class $j$. Then, given a query $\vx \in \cX$ with feature tensor $F \defn \phi_\theta(\vx)$, for each class $j$, a score 
\begin{align}
	s_j(F) \defn 
			\sum_{\vr \in \Omega} \sum_{\vv \in \text{NN}_{V_j}(F(\vr))} \cos(F(\vr), \vv)
\label{eq:nbnn}
\end{align}
is defined as the average cosine similarity over the features $F(\vr)$ at all spatial positions $\vr \in \Omega$ and their $k$-nearest neighbors $\text{NN}_{V_j}(F(\vr))$ in $V_j$. Then, the prediction for $\vx$ is the class of maximum score.

\section{Local propagation}
\label{sec:method}

\subsection{Base training}
\label{sec:base}

\head{Dense classifier.}
We use a \emph{dense classifier} for base training, introduced by Lifchitz \etal~\cite{Lifchitz19}. Rather than flattening or pooling, the classifier $f_{\theta,W}: \cX \to \real_+^{r \times \bc}$ maps an example $\vx$ to a tensor of probabilities over spatial positions, by applying a cosine classifier~\eq{cosine} densely at each position
\begin{align}
	f_{\theta,W}(\vx) \defn \left[
		\vsigma \left( \rho [\cos(\phi_{\theta}(\vx)(\vr), \bvw_j)]_{j=1}^{\bc} \right)
	\right]_{\vr \in \Omega},
\label{eq:dc}
\end{align}
where the class weights $W$ are shared over locations with $\bvw_j \in \real^d$. Cross-entropy applies to all spatial positions using the same class label, and cost function~\eq{ce-cost} becomes
\begin{align}
	J(\bD; \theta, W) \defn
		\sum_{i=1}^n \sum_{\vr \in \Omega} \ell(f_{\theta,W}(\bvx_i)(\vr), \bvy_i).
\label{eq:dc-cost}
\end{align}

\head{Local spatial pooling.}
Dense classification avoids global spatial pooling by going to the other extreme of applying the loss to every position. This happens regardless of whether the effective receptive field is large enough to represent the class at hand, so it assumes an appropriate resolution of the feature tensor. However, it has been observed that it helps to use input images of higher resolution than the standard benchmarks~\cite{dvornik2018}, which we follow. This results in features of accordingly higher resolution, where each position corresponds to small details. We solve this by applying \emph{local spatial pooling} on the feature tensor, both before dense classification at base training as well as at new classification tasks.

\begin{figure}
\setlength\tabcolsep{2pt}
\begin{tabular}{cccccc}
\fig[.15]{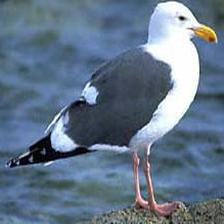} &
\fig[.15]{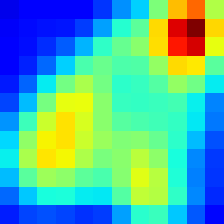} &
\fig[.15]{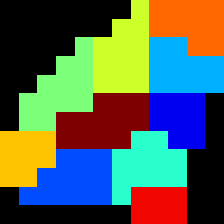} &
\fig[.15]{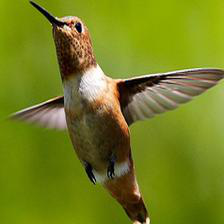} &
\fig[.15]{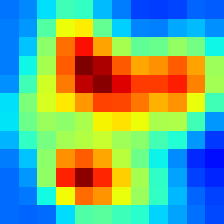} &
\fig[.15]{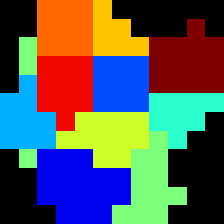} \\
\fig[.15]{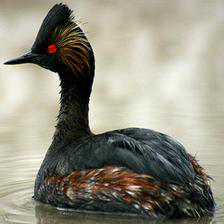} &
\fig[.15]{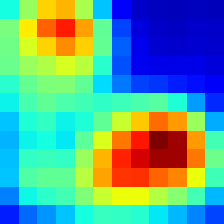} &
\fig[.15]{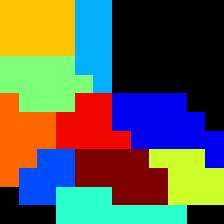} &
\fig[.15]{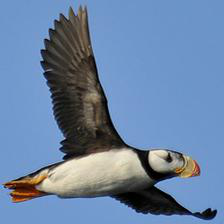} &
\fig[.15]{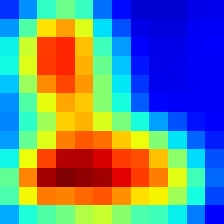} &
\fig[.15]{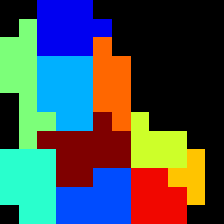} \\
\fig[.15]{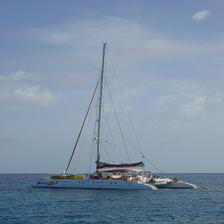} &
\fig[.15]{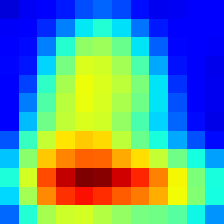} &
\fig[.15]{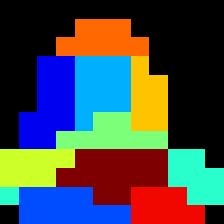} &
\fig[.15]{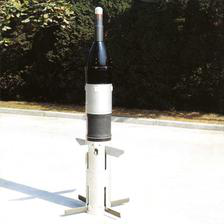} &
\fig[.15]{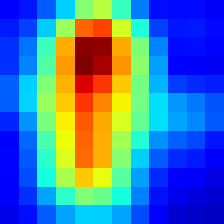} &
\fig[.15]{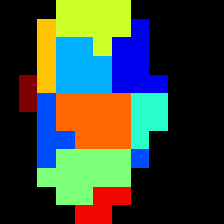} \\
\fig[.15]{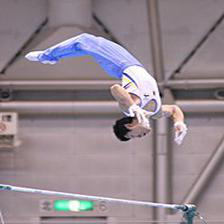} &
\fig[.15]{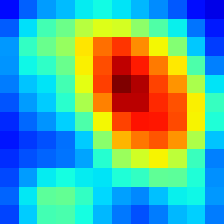} &
\fig[.15]{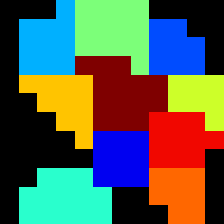} &
\fig[.15]{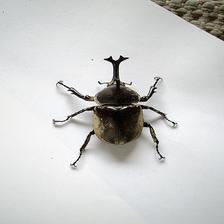} &
\fig[.15]{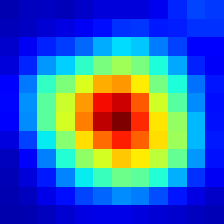} &
\fig[.15]{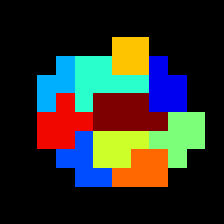}
\end{tabular}
\caption{Examples of images, each with the corresponding spatial attention heatmap and clusters used in feature pooling (black indicates regions below threshold in the heatmap). The first two lines correspond to CUB, the last two to \emph{mini}ImageNet. We use $\tau=0.3$ for spatial attention and $m=10$ for feature pooling.}
\label{fig:ex-att-clust}
\end{figure}

\subsection{Few-shot classification}
\label{sec:novel}

\head{Spatial attention.}
Before we can use features of all spatial positions as data, it is important to suppress the background, which appears frequently across positions and images, without being discriminative for the classification task. There are different approaches, such as learning a class-agnostic \emph{spatial attention} mechanism~\cite{Wertheimer_2019_CVPR,Zhang_2019_CVPR} or simply by a form of pooling over feature channels~\cite{kalantidis2016cross}. We follow the latter approach. In particular, given an example $\vx \in \cX$ with feature tensor $F \defn \phi_{\ntheta}(\vx)$, we select a subset of feature vectors $a(F) \subset \real^d$ at spatial positions $\vr \in \Omega$ where the $\ell_2$-norm is at least $\tau > 0$ relative to the maximum over the domain:
\begin{align}
	a(F) \defn
		\{ F(\vr): \| F(\vr) \| \ge \tau \max_{\vt \in \Omega} \| F(\vt) \|, \vr \in \Omega \}.
\label{eq:norm}
\end{align}
Examples are shown in~\autoref{fig:ex-att-clust}.
We find this mechanism particularly effective for its simplicity, not only for our method, but also for all baselines. No spatial attention is a special case where $\tau = 0$.


\head{Feature pooling.}
Propagation tends to amplify elements that appear frequently in a dataset. Local propagation does the same for elements originating from different spatial positions, which in turn depends on the scale of objects relative to the spatial resolution. This can be particularly harmful with elements originating from background clutter and bypass condition~\eq{norm}, exactly because they appear frequently.

To obtain a fixed-size representation that only depends on the content, we perform pooling in the feature space into a fixed number of vectors per example. We do so by \emph{clustering}: given an example $\vx \in \cX$ with selected feature vectors $a(\phi_{\ntheta}(\vx))$~\eq{norm}, we obtain $m$ clusters by $k$-means. We represent the corresponding $m$ \emph{feature centroids} as columns in the $d \times m$ matrix $g_{\ntheta}(\vx)$.
Examples are shown in~\autoref{fig:ex-att-clust}.
We use this representation only for local propagation. Global propagation and no feature pooling are special cases where $m = 1$ and $m = w \times h$ respectively.


\head{Local propagation.}
We develop this idea under the transductive setting because it is more general: The non-transductive is the special case where $q=1$, the set of queries $Q = \{\vq_1\}$ is singleton and we are making a prediction for $\vq_1$. Given the support examples in $\nD$ and queries $Q$, we represent the feature centroids of both as columns in the $d \times t$ matrix
\begin{align}
	V \defn
		\left(
		\begin{array}{cccccc}
			g_{\ntheta}(\nvx_1) & \dots & g_{\ntheta}(\nvx_n) &
			g_{\ntheta}(\vq_1)  & \dots & g_{\ntheta}(\vq_q)
		\end{array}
		\right)
\label{eq:feat}
\end{align}
where $t \defn (\nn+q)m$. Following~\cite{ITA+17}, we use the pairwise similarity function $s(\vv_1, \vv_2) \defn [\cos(\vv_1, \vv_2)]_+^\gamma$ where $\gamma > 1$, and construct the reciprocal \emph{$k$-nearest neighbor graph} of the columns of $V$, represented by the $t \times t$ symmetric nonnegative \emph{adjacency matrix} $W_V$ with zero diagonal. Following~\cite{ZBL+03}, this matrix is symmetrically normalized as $\cW_V \defn D_V^{-1/2} W_V D_V^{-1/2}$, where $D_V \defn W_V \vone_t$ is the \emph{degree matrix} of the graph and $\vone_t$ is the $t \times 1$ all-ones vector.

Extending~\cite{ZBL+03}, given any matrix $A \in \real^{u \times t}$ (or row vector for $u = 1$), its \emph{propagation} on $V$ is defined as
\begin{align}
	p_V(A) \defn
		A (1-\alpha) (I - \alpha \cW_V)^{-1}.
\label{eq:prop}
\end{align}
This is a smoothing operation on the graph of $V$, where parameter $\alpha \in [0,1)$ controls the amount of smoothing: Columns of $A$ corresponding to similar columns of $V$ are averaged together. It is infinitely-recursive, as revealed by the series expansion of the matrix inverse~\cite{ZBL+03}.

The operation~\eq{prop} is called \emph{local propagation} because the graph is defined on local representations originating from different spatial positions of the given images. \emph{Global propagation} is the special case of having $m = 1$ cluster per image. This is the same as \emph{global average pooling} (GAP), with or without spatial attention.


\head{Local feature propagation.}
Using $A = V$, $u = d$ in~\eq{prop}, \emph{local feature propagation} amounts to propagating $V$ on itself:
\begin{align}
	\wt{V} \defn p_V(V),
\label{eq:fp}
\end{align}
in the sense that similar feature vectors in columns of $V$ are averaged together, becoming even more similar. No feature propagation is a special case where $\alpha = 0$, $\wt{V} = V$.


\begin{figure}
\setlength\tabcolsep{2pt}
\begin{tabular}{cccccc}
\fig[.15]{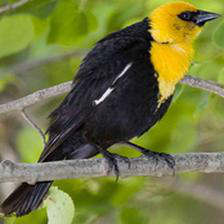} &
\fig[.15]{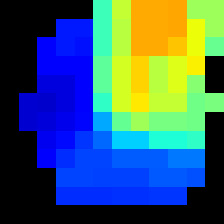} &
\fig[.15]{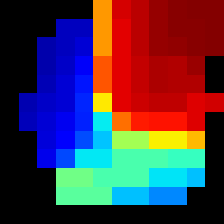} &
\fig[.15]{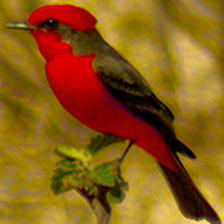} &
\fig[.15]{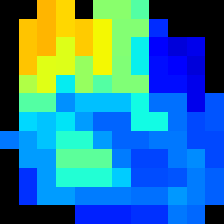} &
\fig[.15]{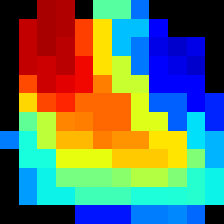} \\
& (a) &&& (b) \\
\fig[.15]{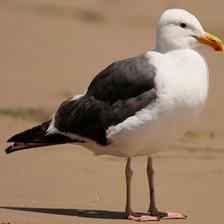} &
\fig[.15]{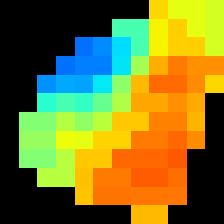} &
\fig[.15]{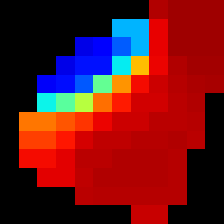} &
\fig[.15]{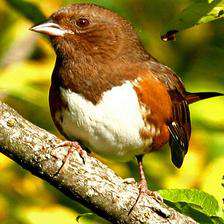} &
\fig[.15]{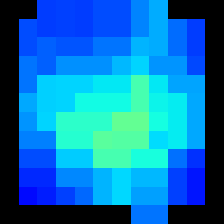} &
\fig[.15]{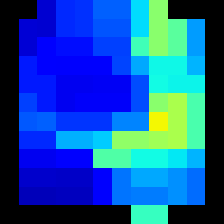} \\
& (c) &&& (d)
\end{tabular}
\caption{Examples of CUB query images in 5-way 5-shot non-transductive tasks, each followed by the heatmap of predicted probability for the correct class using a prototype classifier, then using local label propagation. (a), (b) Local label propagation helps classifying to the correct class. (c) Both give a correct prediction. (d) Local label propagation fails.}
\label{fig:ex-prob}
\end{figure}

\head{Local label propagation.}
Given the propagated features $\wt{V}$~\eq{fp}, we form a new graph with normalized adjacency matrix $\cW_{\wt{V}}$. Extending~\cite{ZBL+03}, we form the $\nc \times t$ zero-one \emph{label matrix} $Y$ with one row per class and one column per spatial position over support examples and queries. A column corresponding to a spatial position of a support example $\vx_i$ is defined as the one-hot label vector $\vy_i$; a column corresponding to a position of a query $\vq_i$ is zero:
\begin{align}
	Y \defn
		\left(
		\begin{array}{cccccc}
			\nvy_1 \vone_m\tran & \dots & \nvy_n \vone_m\tran &
			\vzero_{c \times m}  & \dots & \vzero_{c \times m}
		\end{array}
		\right)
\label{eq:label}
\end{align}
where $\vzero_{c \times m}$ is the $c \times m$ zero matrix and there are $q$ such matrices. Using $A = Y$, $u=c$ in~\eq{prop}, \emph{local label propagation} then amounts to propagating $Y$ on $\wt{V}$:
\begin{align}
	\wt{Y} \defn p_{\wt{V}}(Y),
\label{eq:lp}
\end{align}
such that spatial positions with similar feature vectors obtain similar class scores. This may make little difference on labeled (support) examples, but is a mechanism for spatial positions of \emph{unlabeled} (query) examples to obtain label information as propagated from spatial positions of labeled examples with similar features.


\head{Inference.}
In $\nc \times t$ matrix $\wt{Y}$~\eq{lp}, there is one row per class and one column per spatial position over support examples and queries. $\wt{Y}$ is nonnegative; by column-wise $\ell_1$-normalizing it into $\nc \times t$ matrix $\hat{Y}$, we can interpret columns as probability distributions over classes per position. For each query example $\vq_i$, if $\hat{Y}_i$ is the corresponding $c \times m$ submatrix of $\hat{Y}$, we average these distributions over positions, obtaining a distribution $\vp_i \defn \hat{Y}_i \vone_m / m$. Finally, as in~\eq{pred}, we make a discrete prediction $\pi(\vp_i) = \arg\max_{j \in [\nc]} p_{ij}$ as the class of maximum probability. This operation is similar to NBNN~\eq{nbnn}, but the quantities being averaged have undergone propagation rather than being direct similarities. \autoref{fig:ex-prob} shows examples of predicted probability for the correct class per spatial location. Local label propagation results in spatially smooth predictions that covers a large portion of the object.

\section{Experiments}
\label{sec:exp}

\subsection{Experimental setup}

\head{Datasets.}
We evaluate our method on two datasets that are common in few-shot learning. The first, \emph{Mini}ImageNet, is a subset of ImageNet ILSVRC-12~\cite{russakovsky2014}. It contains 600 images for each of its 100 classes. Following the work of Ravi and Larochelle~\cite{ravi2017}, we use 64 classes for base training, 16 classes for validation and 20 classes for test. We resample all images to 224$\times$224, similarily to~\cite{chen2019,dvornik2018}.
The second dataset, CUB-200-2011~\cite{wah2011}, referred to as CUB below, was introduced to few-shot learning by Hilliard \etal~\cite{hilliard2018}. It contains 11,788 images from 200 distinct bird species. Following the splits of Ye \etal~\cite{ye2018}, we use 100 classes for base training, 50 for validation and 50 for testing. We crop images using bounding box annotations and resample them to 224$\times$224.

\head{Network.}
We test our method on a ResNet-12 embedding network. Introduced in~\cite{oreshkin2018}, this network is now commonly used in the few-shot learning community. With input images of size 224$\times$224, the embedding features are tensors of resolution 14$\times$14. To adapt the the larger images before applying a dense classifier~\cite{Lifchitz19}, we apply average pooling on these feature tensors, with kernel size 3$\times$3 and stride 1 without padding. The resulting tensors are of resolution 12$\times$12.

\head{Base Training.}
We train the network from scratch using stochastic gradient descent with Nesterov momentum on mini-batches of size 32. The learning rate schedule is set according to the 5-way 5-shot validation accuracy.

\head{Evaluation protocol.}
For each dataset, we obtain a unique embedding network resulting from base training. All methods are then applied to the same features. For all experiments, we sample 2000 5-way few-shot tasks from the test set, each with 15 queries per class. We report average accuracy as well as 95\% confidence interval. We evaluate two different settings: In the \emph{non-transductive} setting, queries are treated as 75 distinct sets $Q$ with only one query each, whereas in the \emph{transductive} setting, there is a single set $Q$ with all 75 queries.

\head{Baselines.}
In the \emph{non-transductive} setting, we compare our method with variants of four existing few-shot inference methods. The first, referred to as GAP+Proto, applies \emph{global average pooling} (GAP) on feature tensors and then uses a prototype classifier~\cite{snell2017} on the support set~\eq{proto}. The second is the inference mechanism of the matching network~\cite{vinyals2016}, while the third, referred to as Local Match, is a modified version as follows. For each support example $\nvx$ with feature tensor $F \defn \phi_\theta(\nvx)$, we use local feature vectors $F(\vr)$ at all positions $\vr \in \Omega$ as independent support examples, with the same label as $\nvx$. We do the same on queries and average the class score vectors over positions. The fourth is the inference mechanism of NBNN~\cite{LiWang_2019_CVPR}~\eq{nbnn}. For each method, we experiment with and without our spatial attention mechanism~\eq{norm}. For Local Match and NBNN, we select a subset of local features per image. For GAP+Proto and Matching Net, we apply GAP to the selected subset only.

In the \emph{transductive} setting, we compare with the inference mechanism of global label propagation~\cite{LLP+19,rodriguez2020embedding}, with and without global embedding propagation~\cite{rodriguez2020embedding}. These baselines are again evaluated with and without spatial attention. We always include spatial attention in our local propagation, but we experiment with and without feature pooling, with and without feature propagation.

\begin{figure}
\centering
\extfig{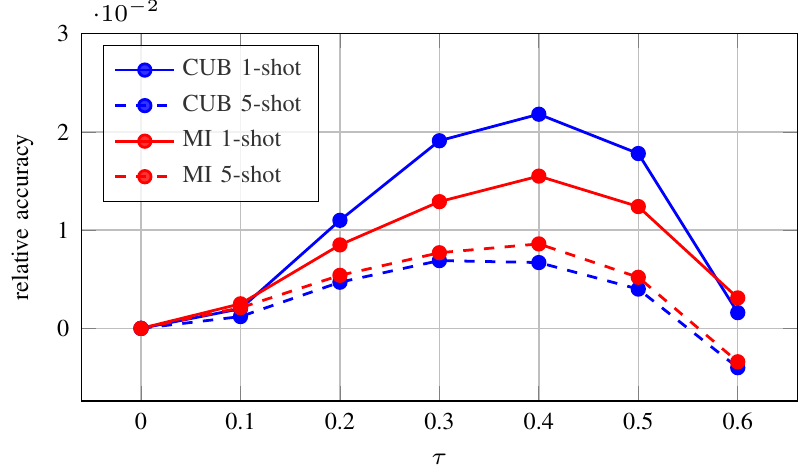}{
\begin{tikzpicture}
\begin{axis}[
	width=\linewidth,
	height=0.6\linewidth,
	font=\scriptsize,
	xlabel={$\tau$},
	ylabel={relative accuracy},
	legend pos=north west,
	ymax = .03,
]
	\addplot[cub,    ] table[x=thres,y expr={\thisrow{GAP1} -.7876}] \attgap; \leg{CUB 1-shot};
	\addplot[cub,five] table[x=thres,y expr={\thisrow{GAP5} -.9184}] \attgap; \leg{CUB 5-shot};
	\addplot[mi,     ] table[x=thres,y expr={\thisrow{MGAP1}-.6519}] \attgap; \leg{MI  1-shot};
	\addplot[mi, five] table[x=thres,y expr={\thisrow{MGAP5}-.8308}] \attgap; \leg{MI  5-shot};
\end{axis}
\end{tikzpicture}
}
\caption{\emph{Spatial attention} on GAP+Proto~\cite{snell2017}: 5-way few-shot classification accuracy \vs threshold $\tau$, relative to $\tau=0$ (no attention).}
\label{fig:att-gap}
\end{figure}
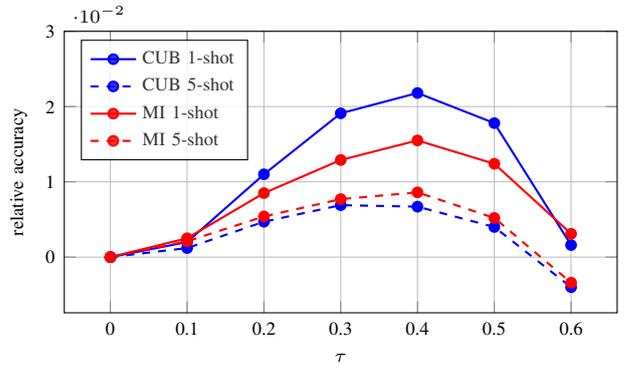

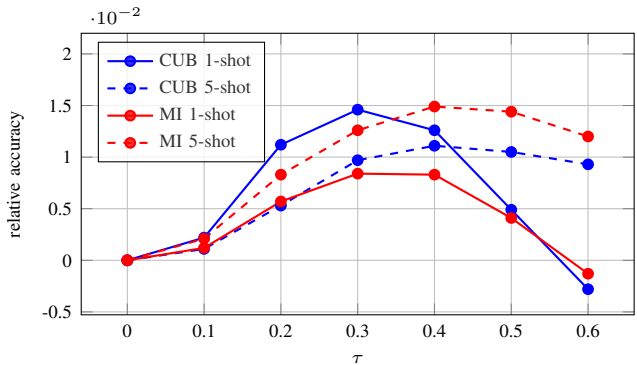
\begin{figure}
\centering
\extfig{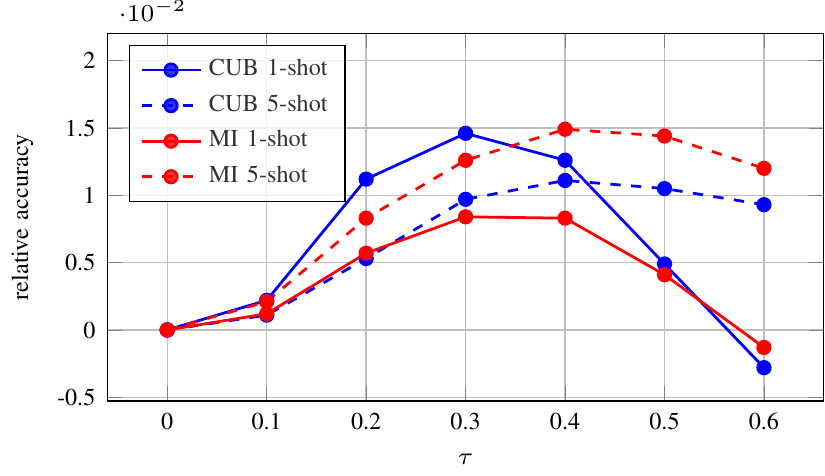}{
\begin{tikzpicture}
\begin{axis}[
	width=\linewidth,
	height=0.6\linewidth,
	font=\scriptsize,
	xlabel={$\tau$},
	ylabel={relative accuracy},
	legend pos=north west,
	ymax=.022,
]

	\addplot[cub,    ] table[x=thres,y expr={\thisrow{CUB1} -.8156}]  \attllp; \leg{CUB 1-shot};
	\addplot[cub,five] table[x=thres,y expr={\thisrow{CUB5} -.9237}]  \attllp; \leg{CUB 5-shot};
	\addplot[mi,     ] table[x=thres,y expr={\thisrow{MI1}  -.6697}]  \attllp; \leg{MI  1-shot};
	\addplot[mi, five] table[x=thres,y expr={\thisrow{MI5}  -.8167}]  \attllp; \leg{MI  5-shot};
\end{axis}
\end{tikzpicture}
}
\caption{\emph{Spatial attention} on our local label propagation, including feature pooling and feature propagation: 5-way few-shot classification accuracy \vs threshold $\tau$, relative to $\tau=0$ (no attention).
All other parameters fixed to optimal.
}
\label{fig:att-llp}
\end{figure}


\subsection{Ablation studies}

Overall, our method has five parameters. Two refer to optional components related to local information: the threshold $\tau$ for spatial attention and the number of clusters $m$ for feature pooling. The other three refer to propagation, like all related methods dating back to~\cite{ZBL+03}: the number of neighbors $k$ in the graph, the exponent $\gamma$ in the feature similarity function and $\alpha$, controlling the amount of the smoothing. For all experiments, we perform a fairly exhaustive parameter search over a small set of possible values per parameter and we make choices according to validation accuracy. We present a summary of parameter search independently for $\tau$ and $m$, keeping other parameters fixed to the optimal.

\head{Spatial Attention.}
As shown in \autoref{fig:att-gap}, referring to GAP+Proto baseline, there is an optimal range of $\tau$ in $[0.3, 0.5]$, such that we filter out the uniformative local feature without removing too much information. The same behavior appears in \autoref{fig:att-llp}, referring to our best method for each setting. For the remaining of the experiments, we fix $\tau$ to 0.3.

\begin{figure}
\centering
\scriptsize
\extfig{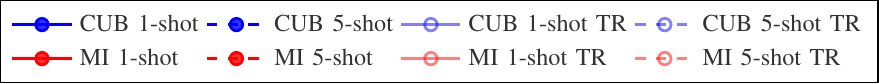}{
\ref*{pool}
}
\extfig{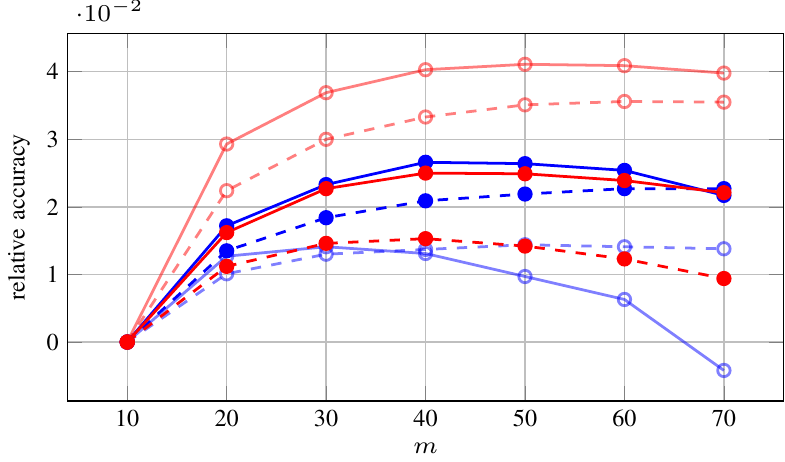}{
\begin{tikzpicture}
\begin{axis}[
	width=\linewidth,
	height=0.6\linewidth,
	font=\scriptsize,
	xlabel={$m$},
	ylabel={relative accuracy},
	legend columns=4,
	legend to name=pool,
]
	\addlegendimage{cub,         }
	\addlegendimage{cub,five     }
	\addlegendimage{cub,     tran}
	\addlegendimage{cub,five,tran}
	\addlegendimage{mi,          }
	\addlegendimage{mi, five     }
	\addlegendimage{mi,      tran}
	\addlegendimage{mi, five,tran}

	\addplot[cub,         ] table[x=clust,y expr={\thisrow{CUB1} -.8048}]  \pool; \leg{CUB 1-shot};
	\addplot[cub,five     ] table[x=clust,y expr={\thisrow{CUB5} -.9107}]  \pool; \leg{CUB 5-shot};
	\addplot[cub,     tran] table[x=clust,y expr={\thisrow{TCUB1}-.8854}]  \pool; \leg{CUB 1-shot TR};
	\addplot[cub,five,tran] table[x=clust,y expr={\thisrow{TCUB5}-.9297}]  \pool; \leg{CUB 5-shot TR};
	\addplot[mi,          ] table[x=clust,y expr={\thisrow{MI1}  -.6542}]  \pool; \leg{MI  1-shot};
	\addplot[mi, five     ] table[x=clust,y expr={\thisrow{MI5}  -.8144}]  \pool; \leg{MI  5-shot};
	\addplot[mi,      tran] table[x=clust,y expr={\thisrow{TMI1} -.7097}]  \pool; \leg{MI  1-shot TR};
	\addplot[mi, five,tran] table[x=clust,y expr={\thisrow{TMI5} -.8121}]  \pool; \leg{MI  5-shot TR};
\end{axis}
\end{tikzpicture}
}
\caption{\emph{Feature pooling} on our local propagation: 5-way few-shot classification accuracy \vs number of clusters $m$, relative to $m=10$ for better visualization. TR: transductive. We use spatial attention in all settings and feature propagation only in transductive.
All other parameters fixed to optimal.
}
\label{fig:pool}
\end{figure}
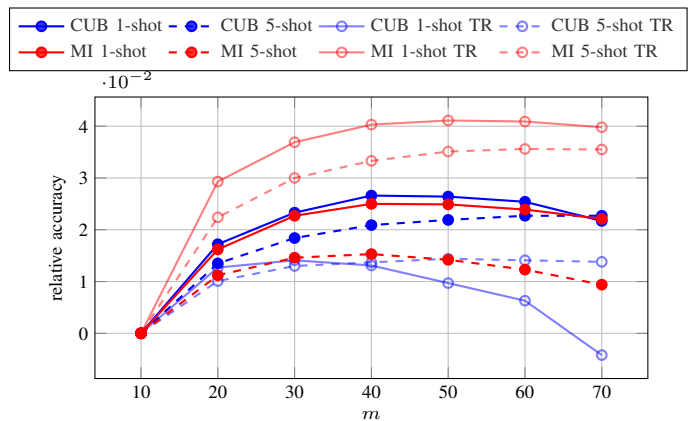

\head{Feature pooling.}
This is a compromise between global pooling and a full set of local features per image, which brings a consistent small improvement compared to both, while making local propagation more efficient by limiting the graph size. According to \autoref{fig:pool}, referring again to our best method for each setting, there is an optimal number $m$ of clusters that depends on the dataset and setting (transductive or not, 1/5-shot). On CUB, we use $m=40$ for 1-shot and $m=60$ for 5-shot. On \emph{mini}ImageNet, we use $m=60$ in both cases.

\head{Propagation parameters.}
Propagation has been extensively researched in the past, so we do not report the study of its parameters. It is known for instance that $\alpha$ should be close to $1$ and there is a local maximum with respect to $k$, which depends on the quantity of the data~\cite{ITA+17}. After parameter search, for most experiments we set $\alpha=0.9$, $\gamma=4$, and $k=5$, $k=50$ respectively for global and local propagation.

\subsection{Results}

\autoref{tab:result} presents a complete set of results our method and baselines in different settings, using different options. We discuss the effect of our contributions below.

\head{Spatial attention.}
We use spatial attention with our method but we also combine it with baselines for fair comparison. It is an extremely simple mechanism that consistently improves few-shot classification accuracy in most cases, including global or local, with propagation or not, transductive or not. The only exception is Local Match on \emph{mini}ImageNet. The gain is more pronounced on 1-shot tasks, which is expected as information selection is more important when information is scarce. It reaches 3\% for the baselines and 2\% for propagation on CUB, as well as 1\% on \emph{mini}ImageNet.

\head{Feature pooling.}
Clustering the set of local features into a given number of clusters for each image is bringing small accuracy improvements when combined with propagation, local or global. In particular, spatial attention and feature pooling brings a 0.30\% to 0.75\% increase of accuracy compared to spatial attention alone on CUB (transductive and non-transductive). An exception is \emph{mini}ImageNet non-transductive where feature pooling gives slightly worse accuracy by an insignificant margin.

\head{Label propagation.}
In the \emph{non-transductive} setting, global label propagation fails. Its performance is similar or inferior to GAP+Protonet. This is to be expected, as this is method a transductive method, so it is not a natural choice given only one query. By contrast, our local label propagation succeeds even in this setting, with up to 2.7\% improvement on CUB 5-way 1-shot compared to GAP+Proto. One exception is \emph{mini}ImageNet 5-way 1-shot, where GAP+Proto is better by a small margin; in this case however, the other local baselines (Local Match and NBNN) are worse than both GAP+Proto and our local label propagation, by a larger margin.

In the \emph{transductive} setting, label propagation, global or local, always helps by using unlabeled data. Our local label propagation with spatial attention and feature pooling improves 5-way 1-shot accuracy over the non-transductive setting by 6\% and 5.5\%, on CUB and \emph{mini}ImageNet respectively. This improvement is lower for 5-shot tasks as more labeled data are used. Compared with global label propagation, it improves by up to 1.5\% on 5-shot, CUB and \emph{mini}ImageNet.

\head{Feature propagation.}
In the \emph{non-transductive} setting, feature propagation is mostly harmful, especially when used with our local label propagation, which remains the best option, together with feature pooling.
In the \emph{transductive} setting however, it helps both global and local label propagation, the only exception being 5-shot, \emph{mini}ImageNet. In the case of local label propagation with feature pooling, the gain is up to 2\% and 1.5\% on 1-shot, CUB and \emph{mini}ImageNet respectively. Therefore this combination is the most effective, improving over our best non transductive result by 8\% and 6.5\% on 1-shot, CUB and \emph{mini}ImageNet respectively.

\head{Universality.}
As shown in \autoref{fig:univ}, our local label propagation is a \emph{universally safe choice} for few-shot inference under both transductive and non-transductive settings. This is in contrast to existing methods such as global label propagation, where the user needs to make decisions depending on the amount of unlabeled data that is available.

\head{Comparison to existing methods.}
\autoref{tab:result} also includes a number of recent few-shot learning methods. For fair comparison, all reported results are using the same ResNet12 as embedding network. We observe that our baseline GAP+Proto is better than these models on non-transductive 5-shot classification on \emph{mini}ImageNet. Our method is then outperforming those models as well. In the transductive setting, global propagation is weaker than existing methods, but our best setting of local propagation (including spatial attention, feature pooling, feature propagation and label propagation) is stronger in general. The only exception is 1-shot classification on CUB, where LR+ICI~\cite{LLP+19} is stronger by a small margin.

In parallel with this work, two methods appeared very recently, which are stronger than our solution on \emph{mini}ImageNet but weaker on CUB: (1) DGPN~\cite{yang2020dpgn}, which is yet another graph-based method and could be easily integrated with our local propagation. (2) DeepEMD~\cite{zhang2020deepemd}, which is based on pairwise image alignment. This is more challenging to integrate, for instance one would need to use alignment in the definition of the graph itself. This can be interesting future work.

\begin{table}
\centering
\scriptsize 
\setlength\tabcolsep{4pt}
\newcommand{\std}[1]{\tiny{$\pm$#1}}
\newcommand{\temp}[1]{{\color{orange}#1}}
\newcommand{\stdmiss}{{\color{white}\std{0.00}}}
\newcommand{\conf}{{\color{black!20!white}\std{0.00}}}
\newcommand{\miss}{{\color{black!20!white}00.00\std{0.00}}}
\begin{tabular}{lccccccc} \toprule
	\mr{2}{\Th{Method}} & \mr{2}{\Th{A}} & \mr{2}{\Th{P}} & \mr{2}{\Th{F}} & \mc{2}{\Th{CUB}} & \mc{2}{\Th{\emph{mini}ImageNet}} \\
	&&&& \Th{1-shot} & \Th{5-shot} & \Th{1-shot} & \Th{5-shot} \\ \midrule
	GAP+Proto~\cite{snell2017}                       &     &     &     & 74.85\std{0.48}        & 90.38\std{0.27}        & 63.39\std{0.46}        & 81.21\std{0.32}        \\
	GAP+Proto~\cite{snell2017}                       & \ch &     &     & 77.10\std{0.47}        & 91.24\std{0.26}        & 64.22\std{0.45}        & 81.71\std{0.31}        \\
	Matching Net~\cite{vinyals2016}                  &     &     &     & 74.85\std{0.48}        & 89.23\std{0.29}        & 63.39\std{0.46}        & 78.14\std{0.33}        \\
	Matching Net~\cite{vinyals2016}                  & \ch &     &     & 77.10\std{0.47}        & 89.95\std{0.28}        & 64.22\std{0.45}        & 78.70\std{0.33}        \\
	Local Match~\cite{vinyals2016}                   &     &     &     & 75.92\std{0.46}        & 89.16\std{0.28}        & 64.05\std{0.46}        & 78.45\std{0.34}        \\
	Local Match~\cite{vinyals2016}                   & \ch &     &     & 78.29\std{0.45}        & 90.60\std{0.26}        & 63.58\std{0.46}        & 78.01\std{0.35}        \\
	NBNN~\cite{LiWang_2019_CVPR}                     &     &     &     & 76.21\std{0.45}        & 89.59\std{0.27}        & 64.90\std{0.45}        & 79.74\std{0.32}        \\
	NBNN~\cite{LiWang_2019_CVPR}                     & \ch &     &     & 79.14\std{0.44}        & 91.40\std{0.25}        & 65.18\std{0.45}        & 80.00\std{0.31}        \\ \midrule
	\mc{8}{\Th{Global Label Propagation, Non-Transductive}}                                                                                                                \\ \midrule
	\mr{3}{Propagation}                              &     &     &     & 74.69\std{0.48}        & 87.96\std{0.30}        & 63.39\std{0.46}        & 75.89\std{0.36}        \\
	                                                 & \ch &     &     & 76.94\std{0.47}        & 89.14\std{0.30}        & 64.22\std{0.45}        & 76.40\std{0.36}        \\
	                                                 & \ch &     & \ch & 77.23\std{0.46}        & 88.78\std{0.31}        & 63.41\std{0.45}        & 77.04\std{0.37}        \\ \midrule
	\mc{8}{\Th{Local Label Propagation (this work), Non-Transductive}}                                                                                                     \\ \midrule
	\mr{4}{Propagation}                              &     &     &     & 78.24\std{0.44}        & 91.07\std{0.26}        & 65.52\std{0.45}        & 80.49\std{0.31}        \\
	                                                 & \ch &     &     & 79.02\std{0.44}        & 91.81\std{0.25}        & \tb{65.74\std{0.45}}   & \tb{81.13\std{0.31}}   \\
	                                                 & \ch & \ch &     & \tb{79.77\std{0.44}}   & \tb{92.07\std{0.25}}   & 65.59\std{0.45}        & 80.73\std{0.31}        \\
	                                                 & \ch & \ch & \ch & 79.32\std{0.44}        & 91.52\std{0.25}        & 64.43\std{0.45}        & 80.26\std{0.32}        \\ \midrule
	\mc{8}{\Th{Global Label Propagation, Transductive}}                                                                                                                    \\ \midrule
	\mr{3}{Propagation}                              &     &     &     & 83.64\std{0.48}        & 90.63\std{0.27}        & 70.07\std{0.51}        & 80.96\std{0.34}        \\
	                                                 & \ch &     &     & 85.52\std{0.46}        & 91.67\std{0.27}        & 70.67\std{0.51}        & 81.44\std{0.33}        \\
	                                                 & \ch &     & \ch & 87.18\std{0.46}        & 91.88\std{0.27}        & 72.54\std{0.54}        & 81.38\std{0.35}        \\ \midrule
	\mc{8}{\Th{Local Label Propagation (this work), Transductive}}                                                                                                         \\ \midrule
	\mr{4}{Propagation}                              &     &     &     & 83.04\std{0.43}        & 91.89\std{0.25}        & 69.95\std{0.48}        & 82.13\std{0.31}        \\
	                                                 & \ch &     &     & 85.33\std{0.42}        & 92.50\std{0.25}        & 71.00\std{0.48}        & \tb{82.87\std{0.30}}   \\
	                                                 & \ch & \ch &     & 85.80\std{0.41}        & 92.92\std{0.24}        & 71.12\std{0.48}        & 82.83\std{0.31}        \\
	                                                 & \ch & \ch & \ch & \tb{87.77\std{0.41}}   & \tb{93.35\std{0.23}}   & \tb{72.57\std{0.51}}   & 82.76\std{0.33}        \\ \midrule
	\mc{8}{\Th{Other models, Non-Transductive}}                                                                                                                            \\ \midrule
	SNAIL~\cite{mishra2018}                          &     &     &     & -                      & -                      & 55.71\std{0.99}        & 68.88\std{0.92}        \\
	TADAM~\cite{oreshkin2018}                        &     &     &     & -                      & -                      & 58.50\std{0.30}        & 76.70\std{0.30}        \\
	DC+IMP~\cite{Lifchitz19}                         &     &     &     & -                      & -                      & 62.53\std{0.19}        & 79.77\std{0.19}        \\
	Neg-Cosine~\cite{liu2020negative}                &     &     &     & -                      & -                      & 62.33\std{0.82}        & 80.94\std{0.59}        \\ \midrule
	\mc{8}{\Th{Other models, Transductive}}                                                                                                                                \\ \midrule
	TPN~\cite{LLP+19}                                &     &     &     & -                      & -                      & 59.46\stdmiss          & 75.65\stdmiss          \\
	LR+ICI~\cite{wang2020instance}                   &     &     &     & 88.06\stdmiss          & 92.53\stdmiss          & 66.80\stdmiss          & 79.26\stdmiss          \\
	EPNet~\cite{rodriguez2020embedding}              &     &     &     & 82.85\std{0.81}        & 91.32\std{0.41}        & 66.50\std{0.89}        & 81.06\std{0.60}        \\ \bottomrule
\end{tabular}
\vspace{3pt}
\caption{5-way few-shot classification accuracy, comparing our local (feature and label) propagation to baselines and existing work. \Th{A}: spatial attention (our work, also applied to baselines). \Th{P}: feature pooling (clustering) (our work). \Th{F}: feature propagation~\cite{rodriguez2020embedding}.}
\label{tab:result}
\end{table}

\section{Conclusion}
\label{sec:conclusion}

Our \emph{local propagation} framework takes the best of both worlds: more data from local representations and better use of this data from propagation. It provides a unified solution that works well given few labeled data and an arbitrary number of unlabeled data. As a result, it works better that solutions meant for the standard few-shot inference and at the same time better than solutions meant for transductive few-shot inference. Two secondary contributions are extremely simple and effective: (a) our \emph{feature pooling} helps control the additional cost related to local features, while improving performance in most cases; (b) our \emph{spatial attention} helps not only our method but all baselines too, by a significant margin on 1-shot classification. Our solution only affects inference, so it can easily be plugged into any alternative representation learning method. It is general enough to integrate other state-of-the-art solutions, like pairwise image alignment, other forms of propagation and propagation on several layers.

\bibliographystyle{IEEEtran}
\bibliography{main}

\end{document}